\documentclass[conference]{IEEEtran}
\IEEEoverridecommandlockouts
\usepackage{cite}
\usepackage{amsmath,amssymb,amsfonts}
\usepackage{algorithmic}
\usepackage{graphicx}
\usepackage{textcomp}
\usepackage{xcolor}
\usepackage{multirow,multicol}
\def\BibTeX{{\rm B\kern-.05em{\sc i\kern-.025em b}\kern-.08em
    T\kern-.1667em\lower.7ex\hbox{E}\kern-.125emX}}
\begin{document}

\title{Improved Image-based Pose Regressor Models for Underwater Environments\\}

\author{\IEEEauthorblockN{Luyuan Peng, Hari Vishnu, Mandar Chitre, Yuen Min Too, Bharath Kalyan and Rajat Mishra}
\IEEEauthorblockA{\textit{Acoustic Research Laboratory, National University of Singapore}}}

\maketitle

\section{Introduction}
Visual localization is a potential solution for the problem of localization in a known underwater environment for inspection. Inspection missions may often involve operations around marine structures, making acoustic navigation with beacons difficult due to shadowing and multipath~\cite{b1}. As the vehicle has to operate close to structures, inertial navigation systems, which accumulate errors with time, may not provide sufficient positioning accuracy~\cite{b1}. In comparison, visual localization using cameras may offer a cost-effective, consistent and accurate alternative in such missions. Previous work has shown that machine learning-based regression methods based on PoseNet\cite{posenet}, can effectively regress a 6-degree-of-freedom (DOF) pose from a single 224$\times$224 RGB image with approximately 6~cm position accuracy and 1.7\textdegree orientation accuracy when tested on simulated underwater datasets~\cite{UwPosenet}. It was also shown that using a deeper neural network as the extractor may improve the model's localization accuracy~\cite{UwPosenet}. This work further investigates the effectiveness of such models on underwater datasets and also explores different techniques to further improve localization performance. We have three main contributions:  
 
 1. We explore the use of long-short-term memory (LSTM)~\cite{lstm_org} in the pose regression model to exploit spatial correlation of the image features and to achieve more structured dimensionality reduction~\cite{lstm}.
 
 2. We test the proposed models on underwater datasets collected from a 1.6~m $\times$ 1~m $\times$ 1~m water-filled tank using a remotely operated vehicle (ROV). The tank offers an environment where we can control lighting and turbidity. The models are able to achieve good accuracy in these datasets, with performance comparable to that obtained with the simulator dataset.
 
 3. The base dataset consist of images taken from the first camera of a stereo camera mounted on the vehicle. Furthermore, we explore the performance improvement obtained by augmenting the data with additional images from the second camera. Fig.~\ref{fig2} shows some examples of underwater scenes in the tank dataset. 

\section{Method}
The pose regression problem is to estimate a 6-DOF pose from a single RGB image. The pose consists of the $x$-$y$-$z$ position and roll-pitch-yaw angle orientation. We use quaternions to represent orientation to avoid wrap-around problems associated with Euler angles~\cite{loss}. Thus, given a monocular RGB image, \emph{I}, the pose-estimator model outputs a 7-dimensional estimated pose vector $\mathbf{y} = [\hat{\mathbf{p}}, \hat{\mathbf{q}}]$ containing a position vector estimate $\hat{\mathbf{p}}$ and orientation vector estimate $\hat{\mathbf{q}}$. For training of the regressing model from images to poses, we use a composite loss function that is a weighted sum of the position error and orientation error squared~\cite{posenet}: 
\begin{equation}
    \mathcal{L} = \mathcal{L}_\mathbf{p}+ \beta \mathcal{L}_\mathbf{q}, \label{eq2}
\end{equation}
where $\mathcal{L}_\mathbf{p} = ||\mathbf{p} - \hat{\mathbf{p}}||_2$ and $\mathcal{L}_\mathbf{q} = ||\mathbf{q} - \hat{\mathbf{q}}||_2$, and $\mathbf{p}$ and $\mathbf{q}$ represent the true pose. $\beta$ is a free parameter that determines the trade-off between desired accuracy in translation and orientation. Based on the requirement of the inspection mission which the estimator is being designed for, we set $\beta$ to 30, which means a position error of 1~m is weighted equivalently to an orientation error of 30\textdegree. 

We implement two different architectures. The first uses pretrained deep convolutional neural networks (DCNN) to extract features from the input images. The resulting feature map is then passed to an affine regressor consisting of dense neural layers to output a 7-dimensional pose vector estimates. The second architecture adds an additional LSTM layer between the DCNN and affine regressor to perform dimensionality reduction by assessing the spatial structure of the image. This is done by four LSTMs which parse the DCNN-output feature map in different directions starting at each of the four image corners to process the spatial trends, and compress it into lower-dimensional information which is easier processed by the affine regressor.

\begin{figure}[t]
\centerline{\includegraphics[width=0.5\textwidth]{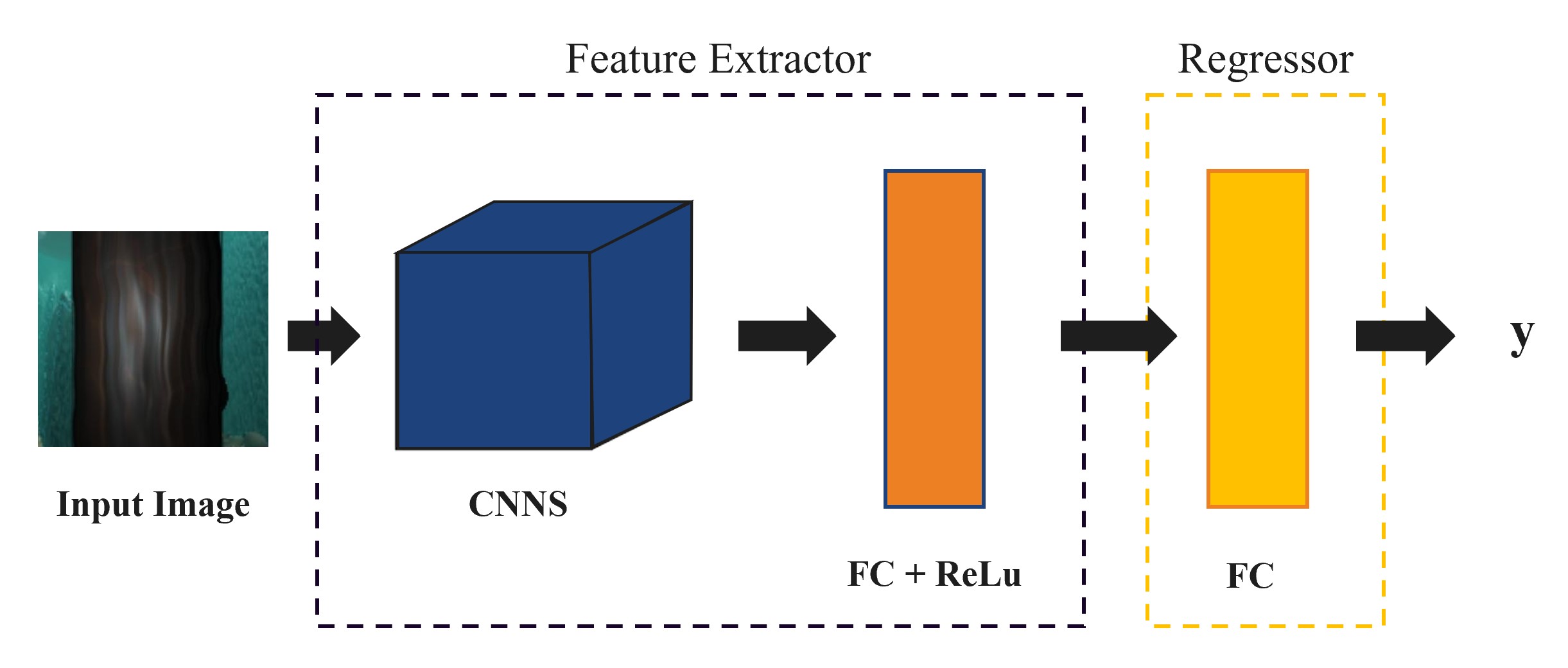}}
\caption{Neural network architecture overview.}
\label{fig2}
\end{figure}

The input images used in the training are rescaled to 256$\times$256 pixels before cropping into a 224$\times$224 feature input using centre cropping. To speed up training, the images are normalized by the mean and standard deviation of the images. The poses are also normalized to lie within [-1, 1].

\section{Datasets}
To train and test our model, we used one dataset collected from an underwater robotics simulator ~\cite{simulator} as shown in Fig.~\ref{fig3} as well as two datasets collected from a tank as shown in Fig.~\ref{fig2}. In the simulator dataset, we operated the ROV in simulation to perform inspection on a vertical pipe in a spiral motion. The total spatial extent covered by the ROV during the inspection is about 2$\times$4$\times$2~m. 14,400 samples of image-pose pair data were collected.  

In the first tank dataset, we operated the ROV in a lawnmower path with translations only (Fig.2), and the rotations were minimal. We collected 3,437 data samples. In the second tank dataset, the ROV primarily performed rotation maneuvers at 5 selected points. We collected 4,977 data samples. We augmented the left-camera dataset by adding the right-camera data, and thereby using the geometry of the stereo camera placement to provide more training data. This worked well and yielded better performance. The total spatial extent covered in tank datasets was 0.4$\times$0.6$\times$0.2~m. 

\begin{table}[b]
\centering
\caption{Mean localization results of several configurations on simulator datasets and tank datasets}
\resizebox{\columnwidth}{!}{%
\begin{tabular}{llll}
Dataset                       & Baseline Model & with ResNet-50 & with LSTM \\ \hline
Simulator & 0.0891~m, 2.91\textdegree & 0.0657~m, 2.15\textdegree & 0.0624~m, 2.06\textdegree   \\
Tank 1  & 0.0427~m, 1.29\textdegree   & 0.0507~m, 1.17\textdegree   & 0.0340~m, 0.786\textdegree           \\
Tank 2  & 0.0464~m, 1.42\textdegree   & 0.107~m, 2.18\textdegree    & 0.0649~m, 1.45\textdegree            \\
Tank 1 (Augmented)  & NA              & 0.0234~m, 4.71\textdegree   & 0.0406~m, 0.568\textdegree          
\end{tabular}%
}
\label{tab1}
\end{table}

\begin{figure}[t]
\includegraphics[width=0.23\textwidth]{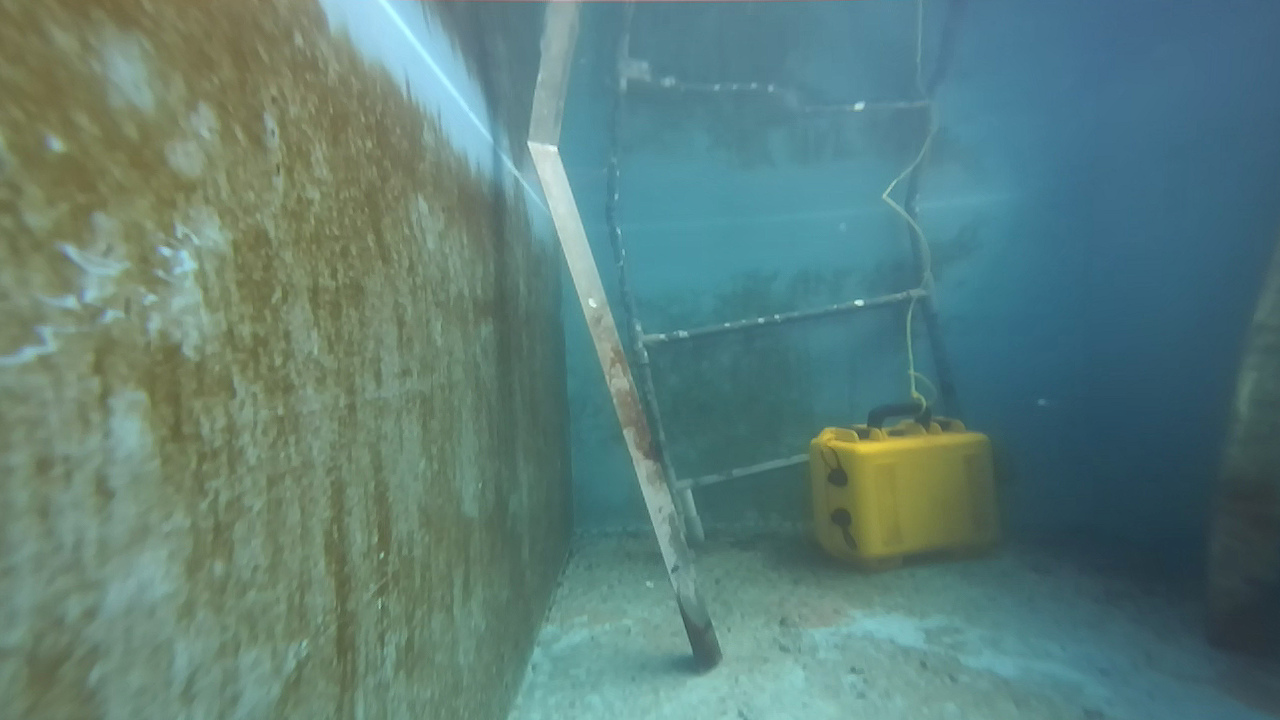} \hfill
\includegraphics[width=0.23\textwidth]{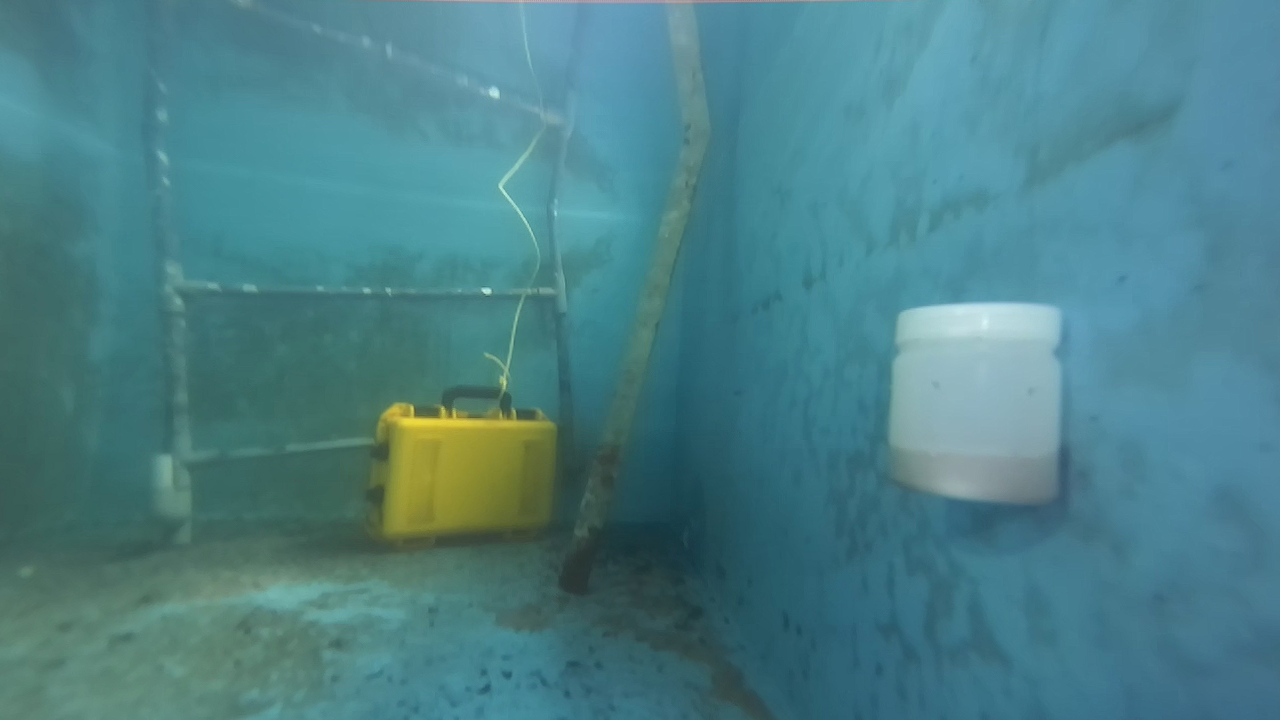}
\caption{Example images from our underwater tank datasets.}
\label{fig2}
\end{figure}

\begin{figure}[t]
\includegraphics[width=0.23\textwidth, height= 0.135\textwidth]{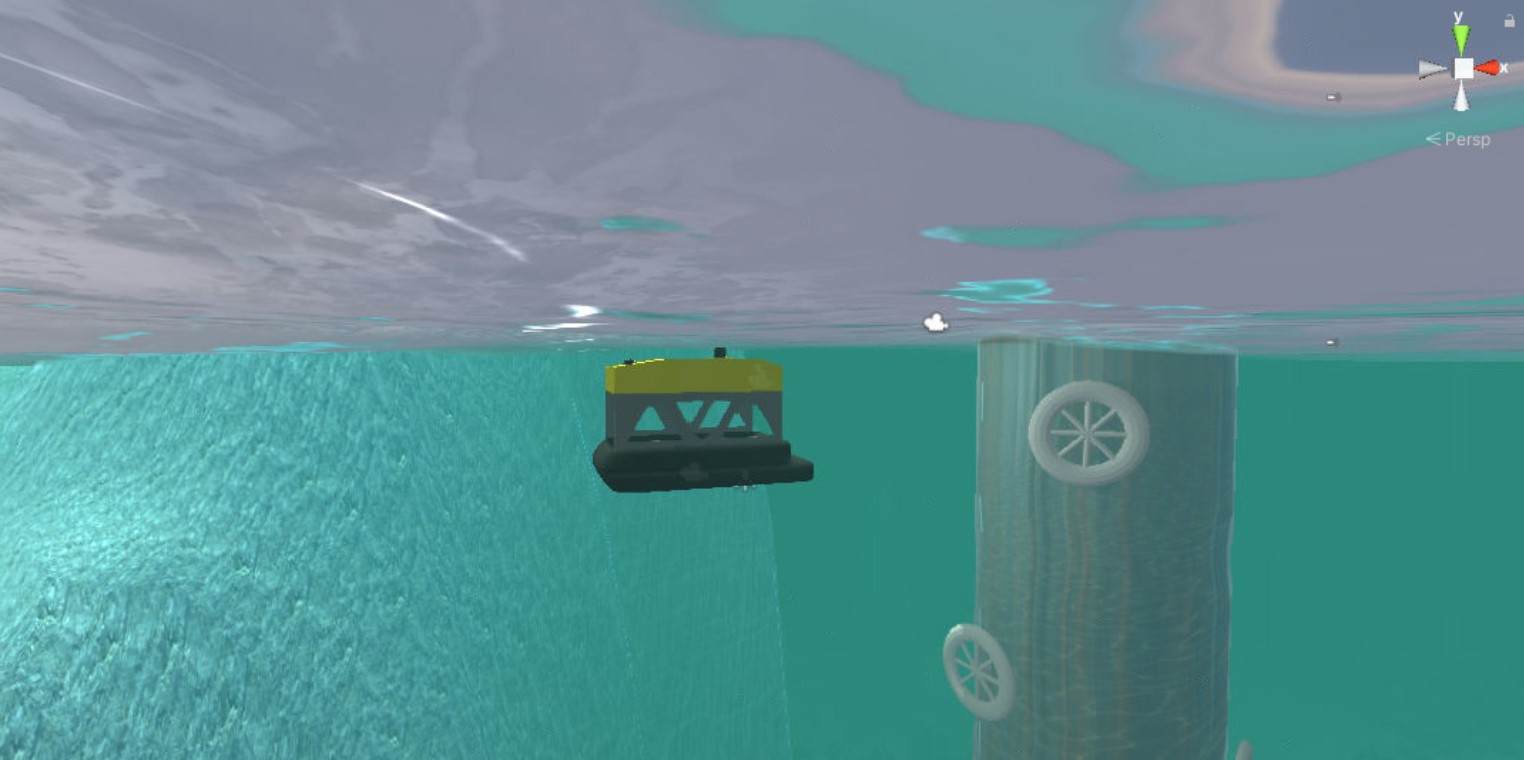} \hfill
\includegraphics[width=0.23\textwidth]{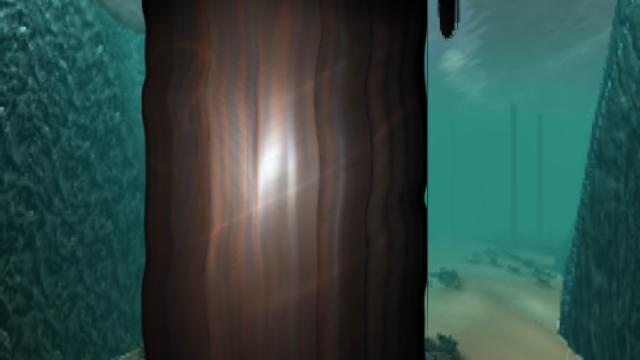}
\caption{Overview of simulated scene in underwater simulator (top) and the simulated image captured by the ROV (bottom).}
\label{fig3}
\end{figure}

\section{Experiments}
We test three model configurations on both the simulator and tank datasets. First, we implement a pose regression model based on GoogLeNet~\cite{googlenet} as our baseline model. Secondly, we implement a model using ResNet-50~\cite{resnet} to evaluate the improvement possible by use of a deeper network. ResNet allows us to have deeper networks with more free parameters, and tries to avoid the problem of overfitting by having residual connections between layers~\cite{resnet}. Thirdly, we implement a pose regression model using an LSTM as an intermediate layer.
\begin{figure}[t]
\includegraphics[width=0.24\textwidth]{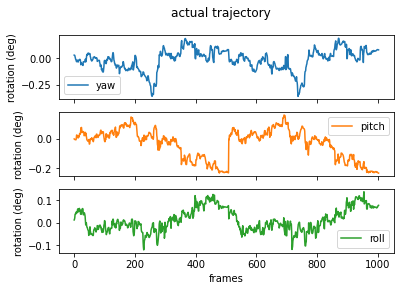}\hfill
\includegraphics[width=0.24\textwidth]{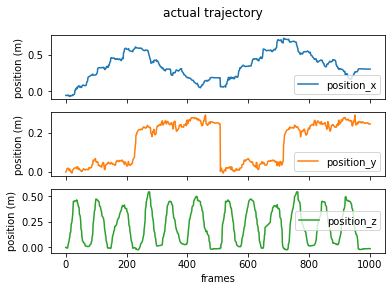}
\\[\smallskipamount]
\includegraphics[width=0.24\textwidth]{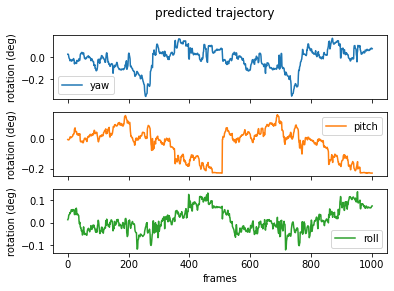}\hfill
\includegraphics[width=0.24\textwidth]{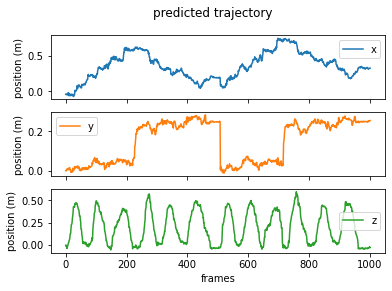}
\caption{\textbf{Predicted Trajectories vs Real Trajectories.} Predicted trajectories with tank datasets 1 and 2 (bottom row) are very close to the actual trajectories (top row).}
\label{fig4}
\end{figure}
The results indicate that all three configurations can perform well in both simulated and tank datasets as shown in Table~\ref{tab1}. The errors are minimal and of comparable magnitude to the noise in the pose recorded by the camera sensors. Fig.~\ref{fig4} shows that the trajectories predicted by the models are very close to the actual trajectories in terms of both position and orientation. With the simulator dataset, which is free of noise, turbidity, light distortion, and other challenges that may be observed in real underwater settings, the ResNet-50 and LSTM-based model performed better than the baseline model. On the other hand, with the tank datasets with some distortions typical of underwater scenarios, ResNet-50 performed worse than the baseline model, whereas the LSTM model did better than baseline with dataset \#1 (which mostly featured translation but almost no rotation). This suggests the ResNet-50 architecture may be slightly overfitting despite the regularization, so it is not able to provide better performance in the tank dataset. We also note that data augmentation significantly improves the model performance, so in cases where there is no significant rotation, the use of data from both cameras can be used to bolster performance. Overall, these methods are robust for application in real underwater environments, and show promise for use in open water settings, where they will be tested next.

\section*{Acknowledgment}
This research project is supported by A*STAR under its RIE2020 Advanced Manufacturing and Engineering (AME) Industry Alignment Fund - Pre-Positioning (IAF-PP) Grant No. A20H8a0241.

\end{document}